\documentclass[10pt,twocolumn,letterpaper]{article}
\usepackage{cvpr}      
\usepackage[accsupp]{axessibility}
\usepackage{float}

\usepackage{tikz}
\usepackage{caption}
\usepackage{multirow}
\usepackage{multirow}

\definecolor{cvprblue}{rgb}{0.21,0.49,0.74}
\usepackage[pagebackref,breaklinks,colorlinks,allcolors=cvprblue]{hyperref}

\def\paperID{39} 
\def\confName{CVPR}
\def\confYear{2026}

\title{F3G-Avatar : Face Focused Full-body Gaussian Avatar}

\author{
Willem Menu\quad
Erkut Akdag \quad
Pedro Quesado \quad
Yasaman Kashefbahrami \quad
Egor Bondarev \\
AIMS Group, Department of Electrical Engineering, Eindhoven University of Technology \\
{\tt\small \{
 w.j.menu, e.akdag,  p.quesado.dos.santos, y.kashefbahrami, e.bondarev\}@tue.nl}
}
\begin{document}
\twocolumn[{%
\maketitle
\begin{center}
  \includegraphics[width=\linewidth]{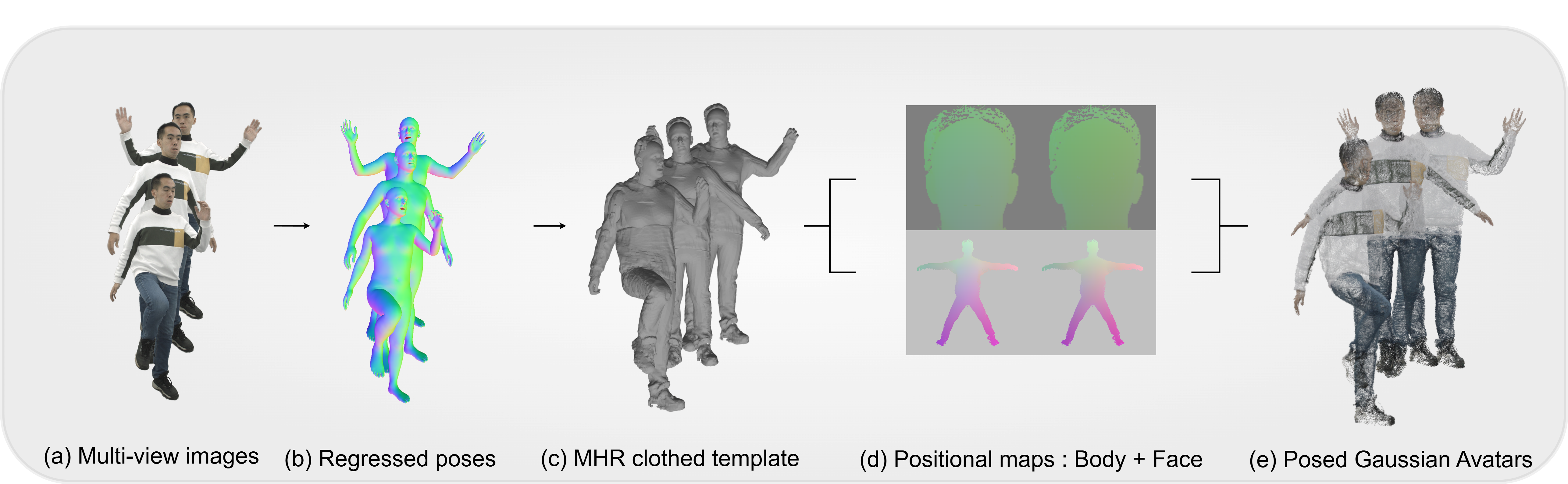}
  \centering
  \captionsetup{type=figure}
  \caption{Framework of F3G-Avatar. Multi-view images and regressed poses are used to generate an MHR clothed template, which is encoded into body and face positional maps and subsequently rendered as posed Gaussian avatars.}
  \label{fig:teaser}
\end{center}
\vspace{0.4em}
}]

\begin{abstract}
\begingroup
\makeatletter
\def\@fptop{0pt}
\makeatother
\endgroup 

Existing full-body Gaussian avatar methods primarily optimize global reconstruction quality and often fail to preserve fine-grained facial geometry and expression details. This challenge arises from limited facial representational capacity that causes difficulties in modeling high-frequency pose-dependent deformations. To address this, we propose F3G-Avatar, a full-body, face-aware avatar synthesis method that reconstructs animatable human representations from multi-view RGB video and regressed pose/shape parameters. Starting from a clothed Momentum Human Rig (MHR) template, front/back positional maps are rendered and decoded into 3D Gaussians through a two-branch architecture: a body branch that captures pose-dependent non-rigid deformations and a face-focused deformation branch that refines head geometry and appearance. The predicted Gaussians are fused, posed with linear blend skinning (LBS), and rendered with differentiable Gaussian splatting. Training combines reconstruction and perceptual objectives with a face-specific adversarial loss to enhance realism in close-up views. Experiments demonstrate strong rendering quality, with face-view performance reaching PSNR/SSIM/LPIPS of 26.243/0.964/0.084 on the AvatarReX dataset. Ablations further highlight contributions of the MHR template and the face-focused deformation. F3G-Avatar provides a practical, high-quality pipeline for realistic, animatable full-body avatar synthesis. The code is available at \url{https://github.com/wjmenu/F3G-avatar}.
\end{abstract}

\section{Introduction}
\label{sec:intro}

Photorealistic, animatable human avatars are the key enabling technology for telepresence, virtual/augmented reality, digital entertainment, and human-computer interaction. The central goal is to capture both the visual appearance and geometric structure of a person in a representation that can be efficiently rendered  from novel viewpoints and driven by motion.

Parametric human body models, most notably the Skinned Multi-Person Linear model (SMPL)~\cite{SMPL:2015} and related variants~\cite{ pavlakos2019expressive, li2017learning}, have become a standard representation for human avatar modeling. They enable recovery of shape, pose, and expression from images or videos through a low-dimensional parameterization of a deformable mesh. SMPL models are animated by adjusting shape and pose parameters and applying linear blend skinning~(LBS)~\cite{anguelov2005scape} to obtain the posed mesh. Many approaches extend these models by incorporating displacement fields to represent clothing, but still struggle with complex geometry and high-frequency detail (e.g., loose garments or fine hair), due to limited topology and texture resolution.

Implicit approaches~\cite{guo2023vid2avatar,li2022tava,peng2021animatable,zheng2022structured}, particularly Neural Radiance Fields (NeRFs)~\cite{mildenhall2021nerf}, model humans as pose-conditioned neural fields learned from RGB videos. However, these methods typically depend on coordinate-based MLPs that are known to suffer from a low-frequency bias. As a result, NeRFs struggle to accurately capture high-frequency details, even when enhanced with learned feature grids or local conditioning. More recently, 3D Gaussian Splatting (3DGS)~\cite{kerbl20233d} has emerged as an efficient explicit alternative, delivering high-quality rendering while significantly improving both the visual quality and rendering speed of prior approaches.

The explicit point-based nature of 3DGS further enables parameterizing appearance and deformation in 2D spaces derived from body template models. This allows the use of powerful 2D backbones for better human avatar modeling. Existing approaches exploit this property by: (i) predicting pose-dependent deformation maps from orthographic front/back projections of a canonical body template~\cite{tao2025impact,li2024animatable}, or (ii) using a 2D parameterization of the underlying human mesh surface in UV space~\cite{jiang2025uv, hu2024gaussianavatar}. In both cases, posed 2D maps are processed with 2D CNNs to predict canonical-space deformations and Gaussian attributes. The obtained Gaussians are then posed via Linear Blend Skinning (LBS) and then visualized by a Gaussian renderer. During training, the Gaussians are optimized to minimize image-based reconstruction losses between the rendered outputs and the corresponding ground-truth camera observations.

Despite achieving strong quantitative performance, these methods are primarily optimized for global, full-body reconstruction and may under-represent important regions that require fine-grained detail. This limitation is most prominent in the face volume, as it occupies only a small fraction of the full-body area. Existing methods tend to allocate insufficient capacity to facial geometry and appearance, leading to oversmoothed features and loss of fine-grained expression detail. However, facial cues play an important role in human perception of identity and realism. When key facial cues are missing or distorted it results in significant perceptual degradation, often associated with the uncanny-valley response~\cite{mori2012uncanny}.

This observation motivates our~\textbf{F3G-Avatar}, a full-body avatar synthesis method that extends the conventional techniques with a dedicated face-focused deformation network. Specifically, a  separate set of canonical Gaussians is generated for the head and the process is driven by additional orthographic front/back projection maps. These maps define a 2D parameter space, where a face-specific deformation network, implemented by StyleUNets~\cite{karras2021alias}, learns high-resolution, pose-dependent Gaussians deformation maps.

To further improve the capture of subtle facial expressions, F3G-Avatar adopts  the Momentum Human Rig (MHR) parametric body model~\cite{ferguson2025mhr}. Compared to commonly-used SMPL-based models, MHR provides more accurate facial articulation due to high-resolution training data and due to sparse, non-linear pose corrective formulation. This leads to improved preservation of local detail and reduces the overly-smoothed or globally entangled deformations observed in conventional parametric models. Furthermore, the coarse garment geometry is modeled on top of the MHR body, enabling consistent deformation of the 3D Gaussians while maintaining alignment during body movements. This yields a clothed parametric template that retains fine-grained control over both facial expressions and body motion. In summary, F3G-Avatar makes the following contributions:


\begin{itemize}
    \item A face-focused canonical deformation network operates alongside the body deformation branch that improves the reconstruction of facial geometry and appearance. The face-focused deformation network independently predicts a set of 3D Gaussians that are concatenated with the Gaussians obtained by the body deformation network.
    \item Integration of the clothed MHR body template into the 3DGS-based avatar method, leading to more accurate reconstruction of facial geometry and expressions. To the best of our knowledge, this is the first implementation of the MHR body model in the context of full-body Gaussian avatar reconstruction.
    \item Comprehensive experimentation that achieves strong performance on AvatarReX and THuman4.0 datasets, with face view PSNR of 26.243/26.934, SSIM of 0.964/0.961, and LPIPS of 0.084/0.062.
\end{itemize} 

\section{Related Work}
\subsection{Parametric Human Body Models}
Conventional human avatar pipelines~\cite{SMPL:2015, pavlakos2019expressive, li2017learning} commonly rely on parametric body models, such as SMPL~\cite{SMPL:2015} or SMPL-X~\cite{pavlakos2019expressive}, which provide representation of human shape and pose through linear blend skinning. The models offer strong priors for articulation, and are widely used for animation, pose estimation, and supervision. However, the fixed topology and limited texture resolution constrain the ability to represent complex geometry, such as loose clothing, fine hair, or subtle view-dependent appearance. As a result, many works augment the models with learned displacement or appearance fields, yet capturing high-frequency detail remains challenging. 

\subsection{Implicit Neural Human Representations}
Implicit approaches~\cite{mildenhall2021nerf, peng2021animatable, li2022tava, guo2023vid2avatar, zheng2022structured} address some of the limitations, by modeling humans as continuous neural fields conditioned on pose. In particular, Neural Radiance Fields(NeRFs)~\cite{mildenhall2021nerf} and animatable extensions learn view-dependent appearance directly from multi-view RGB data. While providing flexibility beyond mesh-based representations, such methods typically rely on coordinate-based MLPs that exhibit a low-frequency bias, limiting the ability to reconstruct fine details. Moreover, volumetric rendering introduces substantial computational overhead, making real-time or high-resolution applications challenging.

\subsection{3D Gaussian-Based Human Avatars}
Recent advances have shifted toward explicit point-based representations, particularly 3D Gaussian Splatting (3DGS)~\cite{kerbl20233d}, which enables efficient rendering with high visual quality. This representation allows modeling deformation and appearance in parameterized 2D space. This 2D parameterization facilitates the use of powerful 2D backbones for predicting pose-dependent Gaussian attributes. A range of approaches build upon this formulation. Animatable Gaussians~\cite{li2024animatable}, predicts pose-conditioned Gaussian maps from orthographic front/back projections. GaussianAvatar~\cite{hu2024gaussianavatar} and 3DGS-Avatar~\cite{qian20243dgs} demonstrate high-quality animatable avatars from monocular or multi-view inputs. SplattingAvatar~\cite{shao2024splattingavatar} stabilizes deformation by embedding Gaussians within a mesh structure. UV-space formulations~\cite{jiang2025uv} exploit surface parameterizations to improve learning stability. Extensions, such as Human Gaussian Splatting~\cite{moreau2024hugs} and HUGS~\cite{kocabas2024hugs}, adapt 3DGS to animatable human modeling under multi-view and monocular settings, while generalizable approaches like HumanSplat~\cite{pan2024humansplat} target single-image reconstruction. Despite these advances, existing methods predominantly optimize for full-body reconstruction quality and tend to distribute model capacity uniformly across the regions of the body. As a result, small yet perceptually critical areas (most notably the face) are often left underrepresented, leading to limited detail and diminished photorealism.

\subsection{Expressive and Perceptual Avatar Modeling}
Head-centered methods allocate model capacity entirely to the face and have consistently advanced facial reconstruction quality. Early approaches combine dynamic NeRFs with morphable face models to enable controllable synthesis and efficient reconstruction. Point-based methods further capture the fine-grained geometric detail through deformable representations \cite{zhao2024psavatar, ploumpis2019combining}, while more recent works integrate 3D Gaussians with parametric face models to achieve precise expression control and high-quality sharp rendering~\cite{gafni2021nerface, zielonka2023instant, zheng2023pointavatar, qian2024gaussianavatars, xu2024gaussian}. Collectively, these studies demonstrate that spatially focused modeling improves facial detail. 
In contrast, full-body methods, such as AvatarRex~\cite{zheng2023avatarrex}, X-Avatars~\cite{shen2023x}, and Expressive Human Avatars~\cite{moon2024exavatar} incorporate expression modeling, but lack a dedicated face-focused deformation mechanism, limiting the ability to fully capture fine-grained facial detail. Perceptual studies indicate that facial appearance plays a dominant role in human judgment of realism~\cite{tao2025impact}. This suggests that full-body systems can benefit from allocating disproportionate capacity to facial detail. 

Motivated by this observation, we introduce a face-focused deformation network alongside the main body deformation network. The face-focused deformation network allows for higher-resolution conditioning and specialized modeling of the facial Gaussians while remaining compatible with full-body rendering. The design reflects an emerging direction toward hybrid representations that combine the efficiency of explicit point-based rendering with region-specific targeting.
\section{Method}
\subsection{3D Gaussian Splatting Preliminaries}
3D Gaussian Splatting (3DGS)~\cite{kerbl20233d} represents a scene as a finite set of anisotropic 3D Gaussian primitives
\begin{equation}
\mathcal{G} = \{G_i\}_{i=1}^{N}.
\end{equation}
Each primitive $G_i \in \mathcal{G}$ is parameterized as
\begin{equation}
G_i = (\mathbf{x}_i, \boldsymbol{\Sigma}_i, \alpha_i, \mathbf{f}_i),
\end{equation}
where $\mathbf{x}_i \in \mathbb{R}^3$ denotes the 3D mean, $\boldsymbol{\Sigma}_i \in \mathbb{R}^{3\times3}$ the covariance matrix, $\alpha_i \in [0,1]$ the opacity, and $\mathbf{f}_i$ the spherical-harmonics coefficients encoding view-dependent color.
\begin{figure*}[t]
    \centering
    \includegraphics[width=\textwidth]{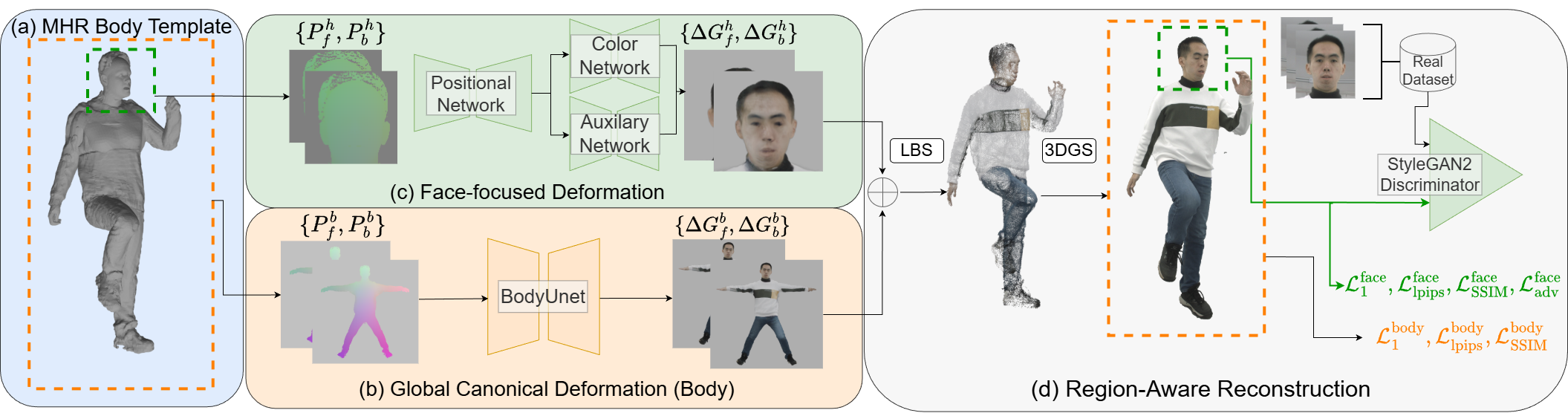}
    \caption{Overview of F3G-Avatar. (a) MHR clothed body template. (b) Global Canonical Deformation (Body): front/back body positional maps are processed by the BodyUNet to predict pose-dependent body Gaussians. (c) Face-focused Deformation: head positional maps drive three StyleUNets to predict positional, color, and auxiliary face attributes. (d) Region-aware Reconstruction : The two branches are fused, posed via LBS, rendered with 3DGS, and optimized with reconstruction losses and a face-specific adversarial loss.}
    \label{fig:pipeline}
\end{figure*}
A 3D Gaussian distribution is defined using the squared Mahalanobis distance
\begin{equation}
d_i^2(\mathbf{p})
=
(\mathbf{p}-\mathbf{x}_i)^\top
\boldsymbol{\Sigma}_i^{-1}
(\mathbf{p}-\mathbf{x}_i),
\end{equation}
such that its density is
\begin{equation}
\mathcal{N}(\mathbf{p} \mid \mathbf{x}_i, \boldsymbol{\Sigma}_i)
=
\frac{1}{(2\pi)^{3/2} |\boldsymbol{\Sigma}_i|^{1/2}}
\exp\left(-\frac{1}{2} d_i^2(\mathbf{p})\right),
\end{equation}
where $\mathbf{p}\in\mathbb{R}^3$ and $|\boldsymbol{\Sigma}_i|$ denotes the determinant of the covariance matrix.

To guarantee that $\boldsymbol{\Sigma}_i$ is symmetric positive semi-definite, it is parameterized via a scale vector $\mathbf{s}_i \in \mathbb{R}^3$ and a unit quaternion $\mathbf{q}_i$:
\begin{equation}
\boldsymbol{\Sigma}_i
=
\mathbf{R}(\mathbf{q}_i)
\operatorname{diag}(\mathbf{s}_i^2)
\mathbf{R}(\mathbf{q}_i)^\top,
\end{equation}
where $\mathbf{R}(\mathbf{q}_i) \in SO(3)$ is the rotation matrix corresponding to $\mathbf{q}_i$.

Given a camera transformation and the Jacobian $\mathbf{J}$ of the projective mapping evaluated at $\mathbf{x}_i$, the covariance is projected into screen space as
\begin{equation}
\boldsymbol{\Sigma}'_i
=
\mathbf{J}\,\boldsymbol{\Sigma}_i\,\mathbf{J}^\top.
\label{eq:sigma_prime}
\end{equation}
For rasterization, only the upper-left $2\times2$ block
\begin{equation}
\boldsymbol{\Sigma}'^{(2D)}_i
=
\boldsymbol{\Sigma}'_{i,1:2,1:2}
\end{equation}
(i.e., the first two rows and columns) is used to define the elliptical footprint in the image plane.

Pixel colors are obtained via front-to-back alpha compositing of depth-sorted Gaussians,
\begin{equation}
C
=
\sum_{i=1}^{N}
\left(
\alpha'_i
\prod_{j=1}^{i-1}(1-\alpha'_j)
\right)c_i,
\label{eq:render}
\end{equation}
where $\alpha'_i$ is the effective opacity at the pixel after evaluating the projected 2D Gaussian and $c_i$ is the view-dependent color obtained from $\mathbf{f}_i$. The parameters of $\mathcal{G}$ are optimized using image-based reconstruction losses, while the number of Gaussians is dynamically adapted through periodic densification and pruning.

\subsection{Overview}
Figure~\ref{fig:pipeline} illustrates the proposed F3G-Avatar method. Given multi-view RGB videos of a subject and the regressed pose and shape parameters, F3G-Avatar reconstructs a realistic representation of both the body and the face. The process starts from a clothed MHR body template, from which the front and back orthographic positional maps are rendered separately for the body and head regions. 
\begin{figure}[b]   
    \centering
    \includegraphics[width=0.45\textwidth]{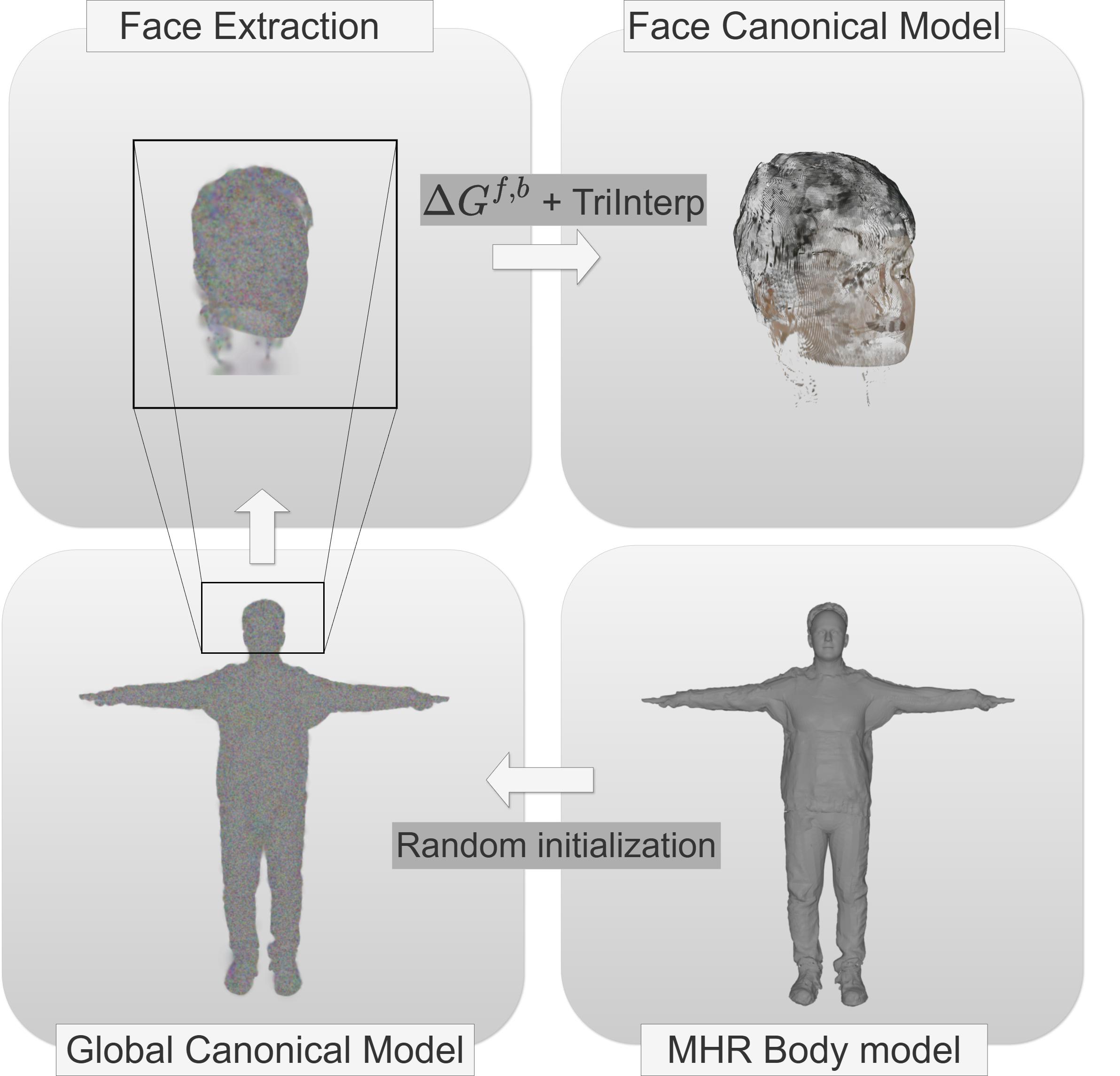}
    \caption{Visualization of the canonical face model construction from the MHR template and head positional maps.}
    \label{fig:init}
\end{figure}
Next, F3G-Avatar is split into two branches: global canonical full-body deformation and face-focused deformation. In the global canonical deformation branch, BodyUNet~\cite{wang2023styleavatar} predicts pose-dependent Gaussian attribute maps in canonical space from the body positional inputs. In parallel, the face-focused deformation branch processes head-specific positional maps using three lightweight StyleGAN-based networks~\cite{karras2021alias}.  The parallel branches predict a set of high-resolution, pose-dependent facial Gaussian maps. The predicted maps define canonical Gaussian primitives for both body and face, which are subsequently deformed and articulated via linear blend skinning (LBS) \cite{anguelov2005scape}. 

Finally, the model is trained with region-aware reconstruction. Full-body reconstruction losses provide global consistency, while additional face-specific perceptual and adversarial losses enhance fine-grained facial detail and realism. 

\subsection{MHR Body Template}
The MHR body template is adopted as the foundation of the representation. Since most multi-view datasets provide SMPL-X parameters (regressed pose and shape parameters), conversion to the MHR representation is required. This conversion is determined by optimizing  
\begin{equation}
    \min_{\beta_{\mathrm{mhr}},\,\theta_{\mathrm{mhr}}}
    \left\lVert V_{\mathrm{mhr}} - V_{\mathrm{smplx}} \right\rVert
    + \lambda \left\lVert J_{\mathrm{mhr}} - J_{\mathrm{smplx}} \right\rVert
\end{equation}
where $\beta_{\mathrm{mhr}}$ and $\theta_{\mathrm{mhr}}$ denote the MHR shape and pose parameters, $V_{\mathrm{mhr}}$ and $V_{\mathrm{smplx}}$ represent the MHR and SMPL-X template vertices, and $J_{\mathrm{mhr}}$ and $J_{\mathrm{smplx}}$ refer to the corresponding joint locations. The reconstruction process starts from a subset of frames in which the subject is captured in a near star-like body pose (A-pose), providing maximal surface visibility across views. From these images, the full clothed-body geometry is reconstructed via implicit surface reconstruction methods~\cite{muller2022instant, yariv2020multiview, li2023neuralangelo, wang2023neus2}, for which NeuS2~\cite{wang2023neus2} is employed. For separation of non-body components (e.g., clothing and accessories), a SAM-based segmentation model~\cite{kirillov2023segany} is applied to the input images. The segmented regions are subsequently projected and attributed onto the reconstructed body mesh using 4D-Dress~\cite{wang20244ddress}. To ensure consistent deformation of these non-body components, Robust Skinning Transfer~\cite{abdrashitov2023robust} is applied to estimate their skinning weights. Finally, the posed MHR model is merged with the segmented non-body components to obtain the complete MHR body template.

The resulting body template is populated with 3D Gaussians, where the positions are based on the vertices of the MHR model in canonical A-pose. The other Gaussian attributes are initialized with informed random values. The canonical 3D Gaussian model is transformed into posed space through LBS. For a canonical Gaussian with position $\mathbf{p}_c$ and covariance $\boldsymbol{\Sigma}_c$, the transformation is given by
\begin{equation}
    \mathbf{p}_p = \mathbf{R}\mathbf{p}_c + \mathbf{t},
    \quad
    \boldsymbol{\Sigma}_p = \mathbf{R}\,\boldsymbol{\Sigma}_c\,\mathbf{R}^\top,
\end{equation}
where $\mathbf{R}$ and $\mathbf{t}$ are the rotation and translation obtained from the Gaussian's skinning weights, and $\mathbf{p}_p$ and $\boldsymbol{\Sigma}_p$ are the Gaussian position and covariance in posed space.
\subsection{Global Canonical Deformation for Body} \label{CanoBody}
A large StyleUNet, $\mathcal{T}(\cdot)$, is employed to capture pose-dependent, non-rigid Gaussian deformations in 2D parameter space. Given the posed MHR templates, front and back position maps $\{P_f^b, P_b^b\}$ are orthographically rendered at a resolution of 1024$\times$1024. Each pixel in the maps corresponds to a single 3D Gaussian with position, covariance, opacity, and color. The maps, together with the camera parameters ($K, R, t$), are fed into the StyleUNet to predict non-rigid deformation maps $\{\Delta G^b_f, \Delta G^b_b\}$. The predicted deformation maps are added to each Gaussian's canonical attributes and then transformed to world space.

In the pretraining stage, StyleUNet is conditioned to reconstruct the input positional maps, while the remaining Gaussian attributes are supervised to match the canonical model. In the subsequent training stage, BodyUNet takes the position maps as input and predicts residual Gaussian attributes that deform the canonical representation. The deformed canonical model is then posed via LBS and rendered.
 
\subsection{Face-Focused Deformation}

\subsubsection{Canonical Face Model}
\label{sec:canonical_face_model}
The canonical face model is initialized by extracting the head region from the pretrained BodyUNet template. After this, Gaussian attributes are estimated by  transforming each head positional map into the canonical frame and averaging across the dataset, which is defined as
\begin{equation}
\mathcal{G}^{v}_{i,j}
=
\frac{1}{N}
\sum_{k=1}^{N}
\mathcal{T}\!\left(
P^{h}_{v,k},\, K_k, \, R_k,\, t_k
\right)_{i,j},
\quad
v \in \{f,b\}.
\end{equation}
Here, $\mathcal{G}^v_{i,j}$ denotes the Gaussian attributes of the head at spatial location $(i,j)$, $v$ indicates if the front or back map is used, $P^h_{v,k}$ represents the $k$-th head positional map in the dataset, and $(K_k, R_k, t_k)$ define the camera calibration parameters for frame $k$.

To increase the face detail, we employ high-resolution positional maps zoomed in on a face. To accommodate the higher spatial resolution of the face positional maps, the canonical Gaussian grid is densified via trilinear interpolation. Formally,
\begin{equation}
\hat{\mathcal{G}}^{v}
=
\operatorname{TriInterp}\!\left(\mathcal{G}^{v}\right),
\end{equation} 
where $\hat{\mathcal{G}}^{v}$ denotes the upsampled canonical face Gaussian representation. Figure~\ref{fig:init} depicts the canonical face model construction. Following the pretraining strategy in \ref{CanoBody}, the face-focused deformation is conditioned on the head region of the pretrained body model. 

\subsubsection{Positional Face Maps}
For face-focused modeling, it is essential to know the precise location of the head within the full-body input images. First, localized crops centered on the face region are extracted. To capture fine-grained facial details, 512$\times$512 crops centered on the face region are extracted. After the crop-and-resize operation, the camera intrinsics must be updated accordingly by
\begin{align}
    f_x' &= s\,f_x, \quad f_y' = s\,f_y, \\
    c_x' &= s\,(c_x - x_c), \quad c_y' = s\,(c_y - y_c),
\end{align}
which yields
\begin{equation}
    \mathbf{K}_{\mathrm{new}} =
    \begin{bmatrix}
        s f_x & 0 & s(c_x-x_c) \\
        0 & s f_y & s(c_y-y_c) \\
        0 & 0 & 1
    \end{bmatrix}.
\end{equation}
Here, $(x_c,y_c)$ denotes the top-left corner of the crop, $(f_x,f_y,c_x,c_y)$ are the original intrinsics, and $(f_x',f_y',c_x',c_y')$ are the updated intrinsics after crop-and-resize. \\
With the updated $K_{\mathrm{new}}$, the MHR posed positional maps are generated using orthographic rendering, resulting in front and back face maps $\{P^h_f, P^h_b\}$.   

\subsubsection{Face-focused Gaussian Maps}
To generate Gaussian maps for the face, we employ three lightweight StyleUNets: Positional ($\mathcal{P}(\cdot)$), Color ($\mathcal{C}(\cdot)$), and Auxiliary ($\mathcal{A}(\cdot)$), which predict positional, color, and auxiliary Gaussian attributes from the head positional maps $\{P^h_f, P^h_b\}$. The positional Gaussian deformation is obtained as
\begin{equation}
\hat{P}^{h}_{v} = \mathcal{P}\!\left(P^{h}_{v}\right), \quad v \in \{f,b\},
\end{equation}
where $\hat{P}^{h}_{v}$ represents the deformed positional map predicted by the positional StyleUNet $\mathcal{P}$. The corresponding color and auxiliary Gaussian attributes are computed using $\mathcal{C}(\cdot)$ and $\mathcal{A}(\cdot)$, respectively. These components are then combined to form the residual Gaussian attribute map:
\begin{equation}
\Delta G^h_v = \mathcal{C}\!\left(\hat{P}^{h}_{v}, K_{\text{new}}, R, t\right)
\mathbin{\|}\, \mathcal{A}\!\left(\hat{P}^{h}_{v}\right)
\mathbin{\|}\, P^{h}_{v}, \quad v \in \{f,b\}.
\end{equation}
Here, $\Delta G^h_v$ denotes the residual Gaussian attribute map for view $v$, corresponding to either the front ($f$) or back ($b$) of the head.

\subsection{Region-Aware Reconstruction}
The predictions are fused with the canonical head Gaussians from Section~\ref{sec:canonical_face_model} and combined with the body Gaussian after Global Canonical Deformation. After combination, the Gaussians are transformed into posed space through LBS. The posed Gaussians are rendered to the image domain using 3D Gaussian splatting (3DGS). To enhance facial detail, a pretrained StyleGAN2~\cite{Karras2021} discriminator is applied to rendered face crops. The discriminator provides a non-saturating adversarial loss, $\mathcal{L}_{adv}$, that is used in addition to reconstruction and perceptual losses.


\begin{table*}[t]
    \centering
    \small  
    \setlength{\tabcolsep}{9.6pt} 
    \renewcommand{\arraystretch}{0.96} 
    \begin{tabular}{l|ccc|ccc}
        \toprule
        & \multicolumn{3}{c|}{Novel View (Body)} & \multicolumn{3}{c}{Novel View (Head)} \\
        \textbf{Model} & PSNR$\uparrow$ & SSIM$\uparrow$ & LPIPS$\downarrow$ & PSNR$\uparrow$ & SSIM$\uparrow$ & LPIPS$\downarrow$ \\
        \midrule
        AvatarRex ~\cite{zheng2023avatarrex}& 23.248 & 0.957 & 0.065 & - & - & - \\
        3DGS-Avatar ~\cite{qian20243dgs} & 28.784 & 0.951 & 0.042 & 24.972 & 0.943 & 0.121 \\
        GaussianAvatar ~\cite{hu2024gaussianavatar} & 26.950 & 0.939 & 0.041 & 24.011 & 0.933 & 0.144 \\
        AnimatableGaussians ~\cite{li2024animatable} & \textbf{30.361} & 0.968 & 0.034 & 25.671 & 0.957 & 0.114 \\
        \midrule
        \textbf{Ours (F3G-Avatar)} & 30.214 & \textbf{0.970} & \textbf{0.032} & \textbf{26.243} & \textbf{0.964} & \textbf{0.084} \\
        \bottomrule
    \end{tabular}
    \caption{Quantitative comparison of full-body and face-focused novel-view synthesis on the AvatarReX~\cite{zheng2023avatarrex} dataset.}
    \label{tab:results_avatarrex}
\end{table*}

\section{Experiments}
\begin{table*}[t]
    \centering
    \small 
    \setlength{\tabcolsep}{9.6pt} 
    \renewcommand{\arraystretch}{1.0} 
    \begin{tabular}{l|ccc|ccc}
        \toprule
        & \multicolumn{3}{c|}{Novel View (Body)} & \multicolumn{3}{c}{Novel View (Head)} \\
        \textbf{Model} & PSNR$\uparrow$ & SSIM$\uparrow$ & LPIPS$\downarrow$ & PSNR$\uparrow$ & SSIM$\uparrow$ & LPIPS$\downarrow$ \\
        \midrule
        TAVA ~\cite{li2022tava} & 26.61 & 0.968 & 0.032 & - & - & - \\
        Ani-NeRF ~\cite{peng2021animatable} & 22.53 & 0.964 & 0.034 & 20.14 & 0.931 & 0.102 \\
        AnimatableGaussians ~\cite{li2024animatable} & \textbf{30.614} & 0.980 & 0.029 & 26.434 & 0.953 & 0.071 \\
        \midrule
        \textbf{Ours (F3G-Avatar)} & 30.311 & \textbf{0.981} & \textbf{0.026} & \textbf{26.934} & \textbf{0.961} & \textbf{0.062} \\
        \bottomrule
    \end{tabular}
    \caption{Quantitative comparison of full-body and face-focused novel-view synthesis on the THuman4.0~\cite{zheng2022structured} dataset.}
    \label{tab:results_thuman}
\end{table*}

\subsection{Evaluation Datasets}

\textbf{AvatarReX}. The AvatarReX~\cite{zheng2023avatarrex} dataset (Real-time Expressive Full-body Avatars) consists of four multi-view human performance sequences captured using 16 synchronized and calibrated RGB cameras arranged in a circular configuration. Each camera records at a resolution of $1500\times 2048$ and 30 fps. For each frame, fitted SMPL-X parameters are provided, supplying pose, shape, and expression estimates. \\

\noindent\textbf{THuman4.0}. Similar to AvatarReX, THuman4.0~\cite{zheng2022structured} provides dense multi-view supervision for animatable human reconstruction. It contains three synchronized sequences captured with 24 calibrated RGB cameras at 30 fps and a resolution of $1330\times 1150$. The dataset includes per-frame SMPL-X registrations.

\begin{figure*}[t]
    \centering
    \includegraphics[width=1\textwidth]{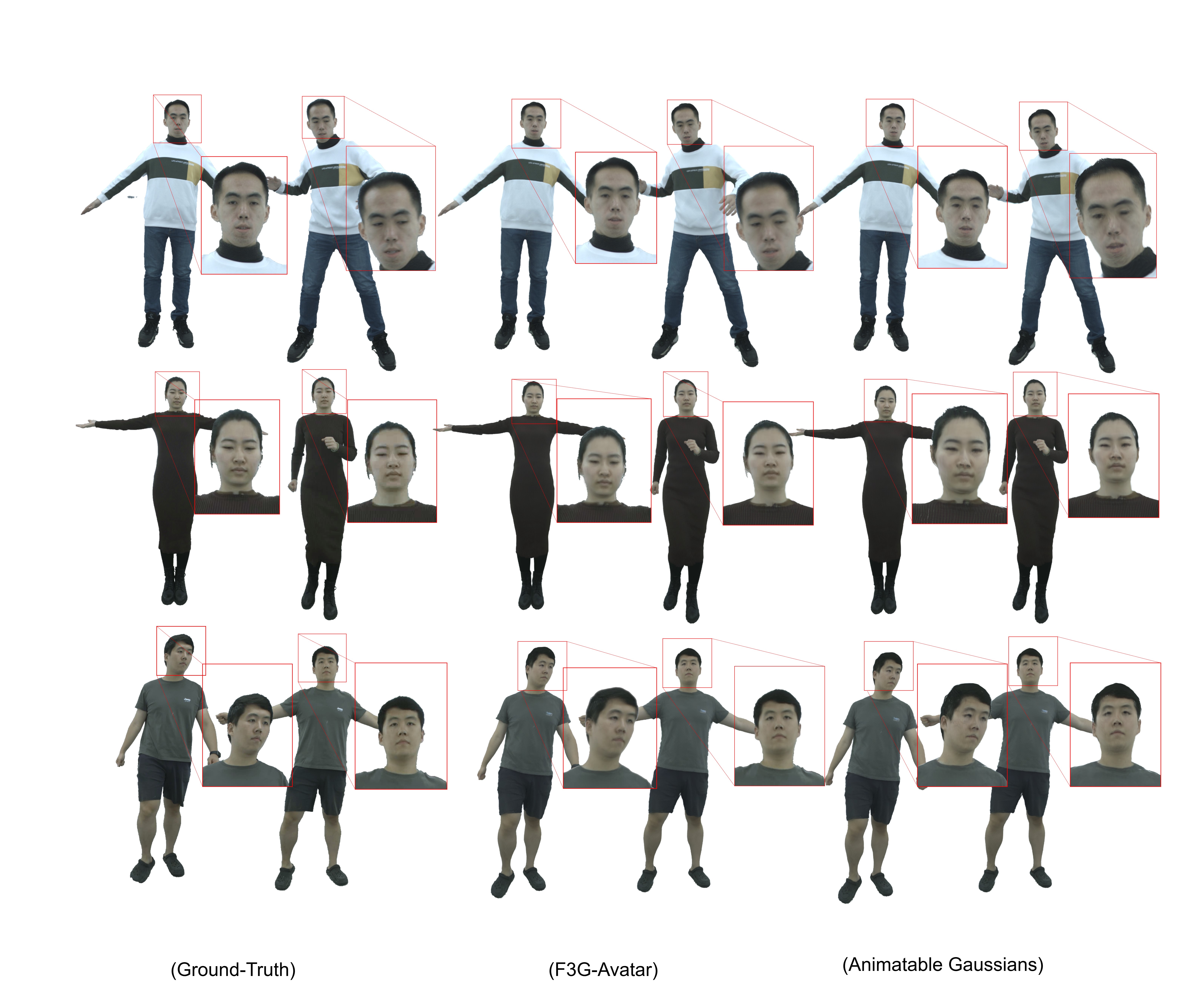}
    \caption{F3G-Avatar displays state-of-the-art rendering quality by delivering improved facial details.}
    \label{fig:qual_views}
\end{figure*}

\subsection{Implementation Details}
\label{sec:impl_details}

\noindent\textbf{Canonical template and Gaussian initialization.}
For each subject, the provided per-frame SMPL-X registrations are used to build a clothed MHR template in a canonical A-pose. 
From the canonical template, front/back concatenated position maps are rendered at a resolution of $1024\times 2048$. The canonical model contains $320\text{k}$ body Gaussians and $60\text{k}$ face Gaussians. The initial centers come from the A-pose position map, with isotropic scales and colors sampled from a uniform distribution. 

\noindent\textbf{Global and Face-focused Deformation architecture.}
The Global Canonical Deformation employs a StyleUNet-based~\cite{karras2021alias} generator to map canonical position maps to Gaussian attributes. 
The backbone processes a $512\times 512$ canonical map and predicts $1024\times 1024$ maps for color, position offsets, and additional Gaussian attributes. 
For face-focused modeling, a lightweight StyleUNet operates on $256\times 256$ face crops to predict head-specific corrections, which are subsequently fused into the global Gaussian representation.

\noindent\textbf{Optimization.} 
At each iteration, a single frame-view is rendered and supervised with RGB and mask. The Global Deformation network is trained on full images, while the Face-focused Deformation network uses cropped head views with updated intrinsics. 
The total loss is a weighted sum of $\ell_1$, LPIPS, and offset regularization term with coefficients $\lambda_{\ell_1}=1.0$, $\lambda_{\text{LPIPS}}=0.1$, $\lambda_{\text{off}}=5\times10^{-3}$, $\lambda_{\text{adv}}=5\times10^{-3}$. 
On AvatarReX, pretraining is performed for $5\text{k}$ iterations, followed by joint optimization for $400\text{k}$ iterations. An additional $5\text{k}$-step face-only fine-tune is applied. 
Training is conducted on a single A100 GPU, requiring approximately 1.5 days per person.

\subsection{Results}

\setlength{\emergencystretch}{1em}
\subsection{Quantitative Results}

Tables~\ref{tab:results_avatarrex} and~\ref{tab:results_thuman} report quantitative comparisons on AvatarReX and THuman4.0 for novel-view synthesis, evaluated with PSNR, SSIM, and LPIPS on both full-body and head regions. On AvatarReX (Table~\ref{tab:results_avatarrex}), F3G-Avatar achieves competitive full-body PSNR (30.214) while obtaining the best SSIM (0.970) and LPIPS (0.032) among the SoTA methods. Although AnimatableGaussians reports a slightly higher PSNR, our approach improves structural similarity and perceptual quality, indicating better preservation of fine details and fewer rendering artifacts. For the head region, F3G-Avatar outperforms prior methods, achieving the highest PSNR (26.243), SSIM (0.964), and a substantially better LPIPS (0.084). On THuman4.0 (Table~\ref{tab:results_thuman}), similar trends are observed. For full-body evaluation, F3G-Avatar achieves the best SSIM (0.981) and LPIPS (0.026), while maintaining PSNR (30.311) competitive to AnimatableGaussians (30.614). For the head region, our method attains the highest PSNR (26.934), SSIM (0.961) and LPIPS (0.062) indicating more accurate facial reconstruction. Overall, the results across both datasets demonstrate that decoupling global and face-specific Gaussian deformations enables improved perceptual quality.
\setlength{\emergencystretch}{0em}

\subsection{Qualitative Results}
Figure~\ref{fig:qual_views} presents qualitative comparisons of rendered avatars on the AvatarReX dataset. We compare F3G-Avatar with AnimatableGaussians under similar novel-view and pose conditions, showing three subjects together with the corresponding ground-truth images. Both methods produce plausible full-body renderings. However, F3G-Avatar consistently preserves sharper facial structures and more stable appearance across viewpoints.
\begin{table}[t]
    \centering
    \small
    \setlength{\tabcolsep}{8pt}
    \renewcommand{\arraystretch}{1.1}
    \resizebox{\columnwidth}{!}{%
    \begin{tabular}{l|ccc}
        \toprule
        Metric & PSNR$\uparrow$ & SSIM$\uparrow$ & LPIPS$\downarrow$ \\
        \midrule
        w/o Face-focused Deformation & 25.721 & 0.951 & 0.107\\
        w/o MHR template & 26.015 & 0.963 & 0.090 \\
        w/o $\mathcal{L_{\text{adv}}}$ & \textbf{27.010} & 0.959 & 0.091\\
        \midrule
        Full model & 26.243 & \textbf{0.964} & \textbf{0.084}  \\
        \bottomrule
    \end{tabular}
    }
    \caption{Ablation study on AvatarReX head novel-view metrics.}
    \label{tab:ablation}
\end{table}

\subsection{Ablation Study}
\textbf{Component ablations.} Table~\ref{tab:ablation} reports face-focused ablations on AvatarReX. Removing the Face-focused Deformation reduces LPIPS and SSIM, while omitting the MHR template lowers PSNR/SSIM. Disabling the adversarial term yields competitive PSNR but worse LPIPS, suggesting the face-specific loss improves perceptual quality.
\textbf{Face-network capacity and input resolution.} Table~\ref{tab:ablation_speed} portrays the effect of both input resolution and face-network capacity. Increasing the input resolution from $128\times128$ to $256\times256$ significantly improves reconstruction quality, raising PSNR from 24.554 to 26.774 and SSIM from 0.939 to 0.956, while reducing LPIPS from 0.101 to 0.086 at the expense of higher runtime. 

\begin{table}[b]
    \centering
    \scriptsize
    \setlength{\tabcolsep}{4pt}
    \renewcommand{\arraystretch}{1.05}
    \resizebox{\columnwidth}{!}{%
    \begin{tabular}{@{}l|cccc@{}}
        \toprule \\
        Metric & PSNR$\uparrow$ & SSIM$\uparrow$ & LPIPS$\downarrow$ & Time (ms)\\
        \midrule
        \textbf{Input Resolution} & & & & \\
        (128 x 128) & 24.554 & 0.939 & 0.101 & 24.34 $\pm$ 1.12 \\
        (256 x 256) & 26.774 & 0.956 & 0.086 & 61.41 $\pm$ 0.87 \\
        \midrule
        \textbf{Model Variations} & & & & \\
        $ (n_{\text{mlp}}, \mathrm{cm})$ : (2, 1) & 25.340 & 0.941 & 0.087 & 47.43$\pm$ 1.11 \\
        $ (n_{\text{mlp}}, \mathrm{cm})$ : (4, 1) & 26.243 & 0.964 & 0.084 & 56.47$\pm$ 1.16\\
        $ (n_{\text{mlp}}, \mathrm{cm})$ : (4, 2) & 26.774 & 0.956 & 0.086 & 61.41$\pm$ 0.87\\
        \bottomrule
    \end{tabular}%
    }
    \caption{Ablation study on model variants with runtime Face-focused Deformation (Time).}
    \label{tab:ablation_speed}
\end{table}

We further vary the StyleGAN-style mapping depth ($n_{\text{mlp}} \in \{2,4\}$) and channel multiplier ($\mathrm{cm} \in \{1,2\}$) in the face sub-networks, while keeping the body network fixed. Increasing the mapping depth from $n_{\text{mlp}}=2$ to $4$ improves reconstruction quality, increasing PSNR and SSIM while slightly lowering LPIPS. Increasing the channel multiplier from 1 to 2 provides a modest PSNR gain (26.24 $\rightarrow$ 26.77). These changes also increase runtime, from 47.43\,ms for the smallest configuration to 61.41\,ms for the widest variation.


\section{Conclusion}
The proposed F3G-Avatar model demonstrates the impact of coupling a clothed canonical template with explicit Gaussian rendering for realistic avatar synthesis. The MHR body template provides a global structure, while pose-conditioned Gaussian deformations capture fine details and maintain view consistency. The body and face branches operate in a complementary manner: the body branch models global non-rigid motion, while the face branch focuses on high-frequency features critical for close-up perception. Quantitative results show consistent improvements across PSNR, SSIM, and LPIPS for body and head views, where F3G-Avatar improves the SoTA results on the AvatarReX and \mbox{THuman4.0} benchmarks. \\
\textbf{Potential Social Impact.} F3G-Avatar can synthesize lifelike, animatable full-body digital humans with realistic facial details, enabling the generation of fabricated 3D content or 2D videos. Therefore, responsible use of this technology is essential.

\section{Acknowledgments}
This work is supported by the ELEVATION Xecs 2023022 project on cloud-based Systems-of-Systems for high-end security and broadcast applications.







{
    \small
    \sloppy
    \bibliographystyle{ieeenat_fullname}
    \bibliography{main}
    \fussy
}


\end{document}